\documentclass[10pt,twocolumn,letterpaper]{article}

\usepackage[pagenumbers]{cvpr}
\usepackage{algorithm}
\usepackage{algpseudocode}
\usepackage{subcaption}
\usepackage{amsmath}
\usepackage{capt-of}
 \usepackage{multirow}
 \usepackage{bbm}
 \usepackage{pdfpages}
\definecolor{cvprblue}{rgb}{0.21,0.49,0.74}
\usepackage[pagebackref,breaklinks,colorlinks,allcolors=cvprblue]{hyperref}

\def\paperID{} 
\def\confName{}
\def\confYear{}

\author{
Yen-Jen Chiou \quad Wei-Tse Cheng \quad Yuan-Fu Yang\\[0.2em]
National Yang Ming Chiao Tung University\\[0.4em]
\texttt{remi.ii13@nycu.edu.tw, andy5552555.ii13@nycu.edu.tw, yfyangd@nycu.edu.tw}}

\title{\textsc{ProFuse}: Efficient Cross-View Context Fusion for Open-Vocabulary 3D Gaussian Splatting}

\begin{document}

\twocolumn[{%
\renewcommand\twocolumn[1][]{#1}%
\maketitle
\begin{center}
    \centering
    \includegraphics[width=1.01\linewidth]{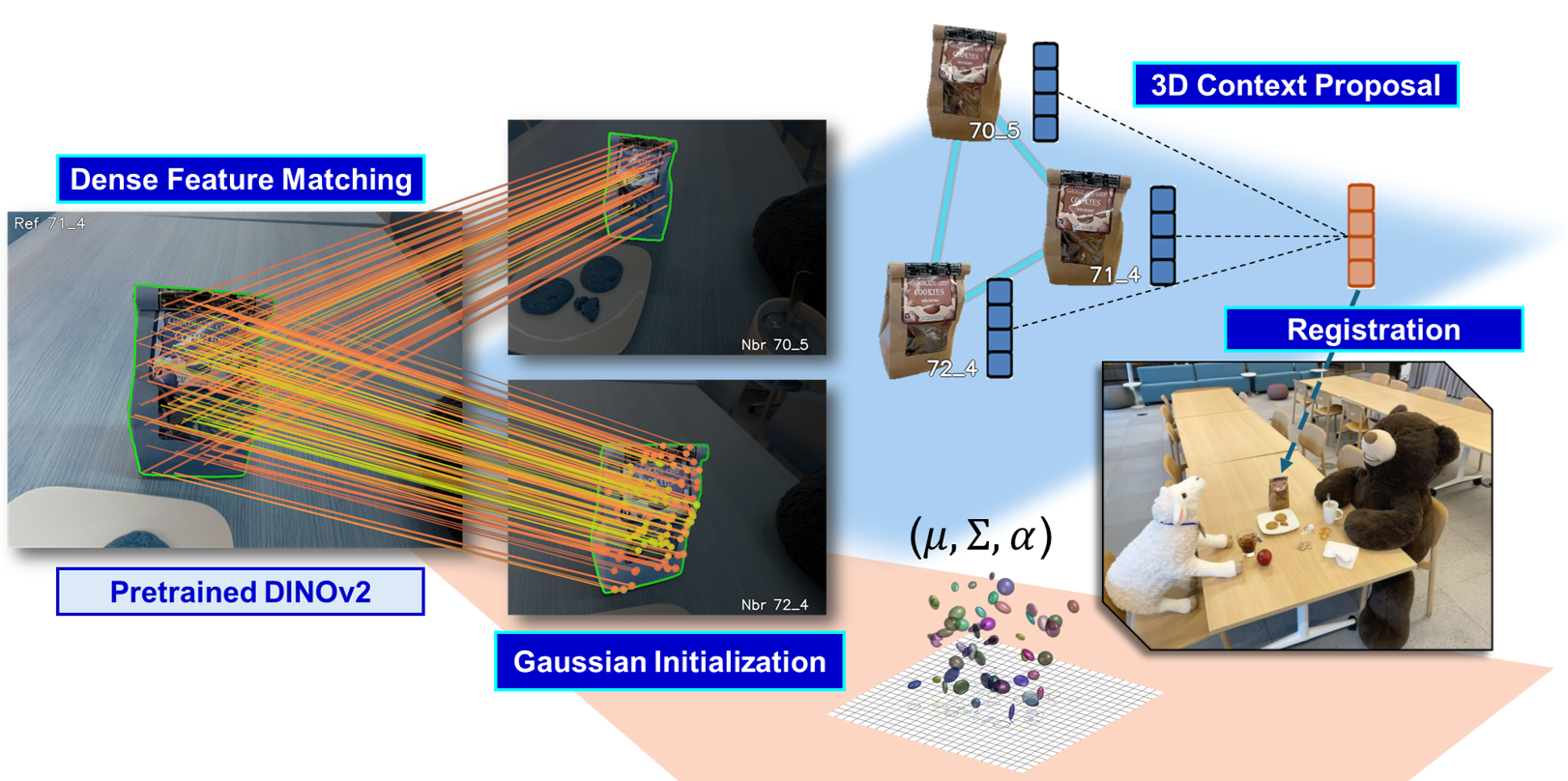}%
    \captionsetup{type=figure}%
    \vspace{0pt}
        \captionof{figure}{Overview of \textbf{ProFuse}. \textbf{Left:} A dense matcher supplies cross-view geometric and semantic correspondences. \textbf{Top:} Warped masks are grouped into 3D Context Proposals with a shared global feature. \textbf{Bottom:} Triangulated matches initialize a compact Gaussian scene, and proposal features are fused without render supervision for coherent open-vocabulary 3D semantics.}
\label{fig:overview}
\end{center}
}]

\maketitle

\begin{abstract}
We present ProFuse, an efficient context-aware framework for open-vocabulary 3D scene understanding with 3D Gaussian Splatting (3DGS). The pipeline enhances cross-view consistency and intra-mask cohesion within a direct registration setup, adding minimal overhead and requiring no render-supervised fine-tuning. Instead of relying on a pretrained 3DGS scene, we introduce a dense correspondence–guided pre-registration phase that initializes Gaussians with accurate geometry while jointly constructing 3D Context Proposals via cross-view clustering. Each proposal carries a global feature obtained through weighted aggregation of member embeddings, and this feature is fused onto Gaussians during direct registration to maintain per-primitive language coherence across views. With associations established in advance, semantic fusion requires no additional optimization beyond standard reconstruction, and the model retains geometric refinement without densification. ProFuse achieves strong open-vocabulary 3DGS understanding while completing semantic attachment in about five minutes per scene, which is 2× faster than SOTA. Additional details are available at our project page \href{https://chiou1203.github.io/ProFuse/}{https://chiou1203.github.io/ProFuse/}.
\end{abstract}

\section{Introduction}
Open-vocabulary 3D scene understanding aims to understand a physical scene using free-form natural language queries, with applications ranging from robotics and autonomous navigation to augmented reality \cite{lerftogo2023, kashu2023openfusion, huang23vlmaps, zhai2024splatloc3dgaussiansplattingbased, chen2025taoavatarrealtimelifelikefullbody, yan2024gsslamdensevisualslam}. The task remains challenging, as the system must recover accurate geometry while also assigning meaningful semantic concepts without being restricted to fixed labels. Earlier efforts explored a range of 3D representations \cite{lerf2023, shen2023distilledfeaturefieldsenable, Peng2023OpenScene, kashu2023openfusion, conceptfusion, nguyen2023open3dis, he2024unimov3dunimodalityopenvocabulary3d}. Recent work has focused on 3D Gaussian Splatting \cite{kerbl3Dgaussians}, which represents a scene as a set of anisotropic Gaussians and enables photo-realistic, real-time rendering.

Early work adopts 2D vision–language distillation in which images are rendered during training and Gaussian features are optimized to match 2D predictions \cite{shi2023language,qin2024langsplat, guo2024semantic, zhou2024feature, gaussian_grouping}. This pipeline can propagate open-vocabulary knowledge into 3D, but it also introduces two structural issues. The supervision signal is delivered only after rendering and compositing, leading to mismatches with the original language embedding that described the region. In addition, semantics are acquired and queried through individual views, making reasoning less direct and less stable. These limitations have motivated methods that operate directly in 3D Gaussian space \cite{wu2024opengaussian, drsplat25,goi2024,li2024instancegaussianappearancesemanticjointgaussian}. These approaches assign language features to each Gaussian and answer a text query by comparing the query embedding with those per-Gaussian features in 3D. 

More recent work has moved toward a registration-based formulation \cite{drsplat25}. This approach bypasses render-supervised semantic training. Language-aligned features are directly registered in Gaussians using their visibility along each viewing ray. The result is a compact, queryable 3D semantic field with high efficiency. Despite such progress, the direct registration paradigm is still in its early stages. Our aim is to strengthen the registration framework by injecting semantic consistency into the 3DGS representation without any additional render-supervised training.

We propose a registration-based framework ProFuse that strengthens semantic coherence in 3D Gaussian Splatting. Our key insight is to enforce two key factors highlighted by previous work \cite{sun2025cagsopenvocabulary3dscene,gaussian_grouping,shen2025trace3dconsistentsegmentationlifting,wu2024opengaussian}, namely cross-view consistency and intra-mask cohesion. Prior approaches typically encourage these properties through render-supervised training on 2D feature maps or through explicit feature-learning objectives. The registration pipeline does not impose these constraints. Our approach injects these forms of semantic consistency directly into the registration framework.

An overview of the proposed pipeline is shown in Figure~\ref{fig:overview}. We introduce a pre-registration stage guided by dense multi-view correspondence \cite{edstedt2023romarobustdensefeature}. The correspondence signal initializes the 3D Gaussian scene with accurate geometry \cite{kotovenko2025edgseliminatingdensificationefficient}, which allows the representation to cover the scene without relying on iterative densification. The same signal is also used to connect observations of the same object across different viewpoints, consolidating them into consistent, object-level groups that we refer to as 3D Context Proposals. Each 3D Context Proposal encodes an object as it appears across views, rather than as an isolated per-frame mask, and provides a stable source of semantics that is aligned across viewpoints.

During feature registration, each proposal carries a global language feature computed from its mask members. We then assign each Gaussian to its corresponding context proposals and associate the global semantics to the Gaussian. Notably, our method does not involve gradient-based fine-tuning or backpropagation of language loss. Through experiments across open-vocabulary 3D perception tasks, we demonstrate effectiveness in 3D object selection, open-vocabulary point cloud understanding, and optimizing efficiency. Our contributions are summarized as follows:
\begin{itemize}[leftmargin=*,noitemsep,topsep=0pt]
\item A registration-based semantic augmentation of 3D Gaussian Splatting that introduces cross-view semantic consistency and intra-mask coherence without any render-supervised training for semantics.
\item A pre-registration stage driven by dense multi-view correspondence. The same correspondence signal initializes a well-covered 3D Gaussian scene and assembles consistent mask evidence across views into 3D Context Proposals. 
\item A unified open-vocabulary 3D scene representation that improves object selection, point cloud understanding, and training efficiency on existing benchmarks while maintaining render-free semantic association efficiently. 
\end{itemize}

Overall, ProFuse offers a compact and training-free route to consistent open-vocabulary 3D scene understanding built directly on correspondence-driven registration.

\begin{figure*}[t]
    \centering
    \includegraphics[width=1.0\linewidth]{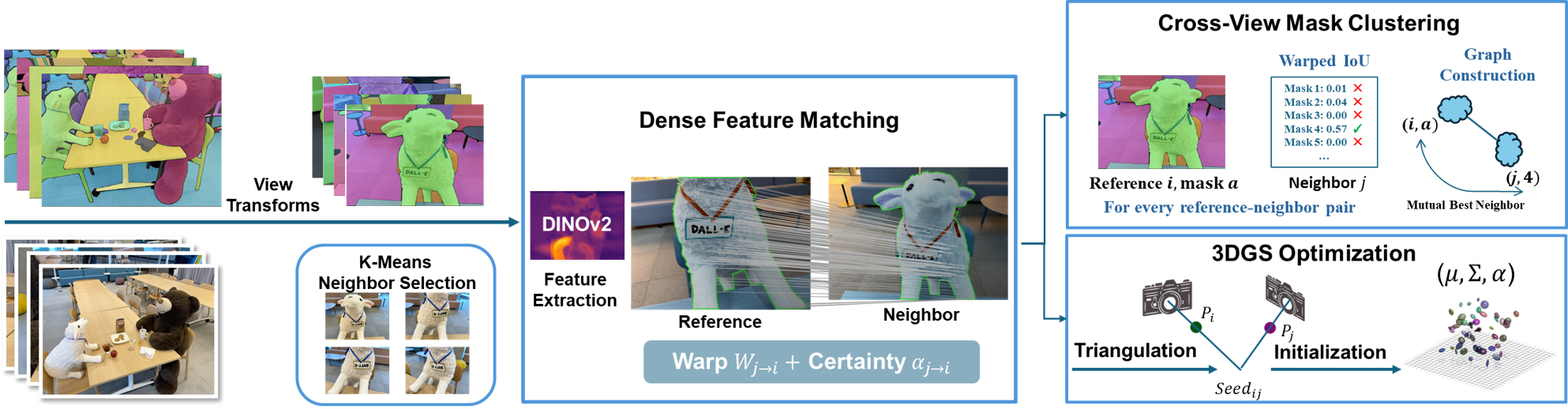}%
    \vspace{-2mm}
    \caption{\textbf{Pre-registration.}
For each reference view we select $K$ neighbors via view clustering, then apply a pre-trained dense matcher to obtain per-pixel warps $W_{j\!\to i}$ and confidences $\alpha_{j\!\to i}$.
\textbf{Bottom right:} Given the warps of a \emph{pixel pair}, we triangulate a 3D seed point for Gaussian initialization.
\textbf{Top right:} Warped IoU comparison on every reference–neighbor \emph{mask pair}; masks that pass the selection form edges of a bipartite graph.}
    \label{fig:pre-registration}
    \vspace{-3mm}
\end{figure*}

\section{Related Work}
Neural rendering has progressed from NeRFs to explicit point-based primitives \cite{mildenhall2020nerf,barron2021mipnerf,mueller2022instant}. 3DGS provides fast, spatially local rendering and is now a common backbone for open-vocabulary understanding \cite{kerbl3Dgaussians,szymanowicz24splatter}.
Render-supervised distillation methods transfer 2D vision-language signals into 3D by supervising rendered feature maps \cite{lerf2023,Peng2023OpenScene,conceptfusion,shi2023language,qin2024langsplat,zhou2024feature,guo2024semantic,sun2025cagsopenvocabulary3dscene,takmaz2023openmask3dopenvocabulary3dinstance,radford2021learningtransferablevisualmodels}.
Direct 3D retrieval attaches language-aligned descriptors to Gaussians or points for volumetric querying \cite{wu2024opengaussian,drsplat25,goi2024,li2024instancegaussianappearancesemanticjointgaussian}.
To stabilize semantics across views, recent works encourage cross-view consistency and semantic cohesion \cite{wu2024opengaussian,chacko2025liftinggaussianssimplefast,piekenbrinck2025opensplat3dopenvocabulary3dinstance,cen2025segment3dgaussians,li2024instancegaussianappearancesemanticjointgaussian,sun2025cagsopenvocabulary3dscene,gaussian_grouping,kundu2022panopticneuralfieldssemantic,Peng2023OpenScene,conceptfusion,takmaz2023openmask3dopenvocabulary3dinstance}.
Finally, dense correspondence provides wide-baseline matches and confidences useful for multi-view grouping and correspondence-driven 3DGS initialization \cite{edstedt2023romarobustdensefeature,cao2024loflatlocalfeaturematching,edstedt2022dkmdensekernelizedfeature,dust3r_cvpr24,mast3r_eccv24,sarlin2020supergluelearningfeaturematching,kotovenko2025edgseliminatingdensificationefficient}. We build on this direction to couple correspondence-guided context association with registration-based semantic field.

\section{Method}

We construct a semantic 3D Gaussian scene that can be queried with natural language without any render-supervised semantic training. The pipeline begins with a pre-registration stage via dense correspondence. This stage initializes a dense Gaussian scene and links segmentation masks across views to form 3D Context Proposals. Each proposal records which masks across views are inferred to refer to the same scene content, giving us cross-view groupings before any semantic fusion. A context-guided registration stage then uses these proposals to compute a global language feature for each proposal. The features are then assigned to the corresponding Gaussians using visibility-based weights derived from transmittance and opacity along camera rays. The final output is a 3D representation with cross-view consistency and intra-mask cohesion that can be searched directly in 3D by a text query.

\subsection{Dense Correspondence Pre-registration}

The pre-registration process begins from a set of posed RGB images of a scene. Let \( \{ I_i \}_{i=1}^{N} \) denote input views, and let each image \( I_i \) have known camera intrinsics and extrinsics. The goal of this stage is to initialize a dense set of 3D Gaussians with accurate geometry and initial appearance attributes, and to record cross-view evidence for semantic grouping. As an overview, the full pre-registration workflow is visualized in Figure~\ref{fig:pre-registration}.

For each image \( I_i \), we obtain a set of non-overlapping region masks \( \{ M_i^k \} \) using SAM \cite{kirillov2023segment}, where \( M_i^k \in \{0,1\}^{H \times W} \) is a binary mask for the region \( k \) in view \( i \). For every mask \( M_i^k \), we extract a language-aligned feature vector \( f_i^k \in \mathbb{R}^{D} \) by cropping the corresponding region in \( I_i \) and encoding it with CLIP \cite{radford2021learningtransferablevisualmodels}. The result is a per-view dictionary
\(\mathcal{S}_i = \{ (M_i^k, f_i^k) \,\mid\, k = 1, \dots, K_i \},\)
where \( K_i \) is the number of predicted regions in view \( i \). The sets \( \mathcal{S}_i \) will later serve as semantic evidence.

\paragraph{Dense Feature Matching.}
To relate content across views, we compute dense correspondences between pairs of images using a pretrained dense matching network (see Figure~\ref{fig:pre-registration})
. The network was trained on a coarse layer using DINOv2 \cite{oquab2024dinov2learningrobustvisual} and a fine layer with pyramid convolution. The result is a robust dense feature matching. 

Given two images \( I_i \) and \( I_j \), the dense matcher returns
\(C(I_i, I_j) \rightarrow W_{j \rightarrow i}, \ \alpha_{j \rightarrow i},\)
where \( W_{j \rightarrow i} \in \mathbb{R}^{2 \times H \times W} \) is a dense warp field that maps each pixel coordinate \( (u,v) \) in \( I_j \) to a subpixel coordinate in \( I_i \), and \( \alpha_{j \rightarrow i} \in \mathbb{R}^{H \times W} \) is a confidence map. Intuitively, \( W_{j \rightarrow i}(u,v) \) predicts where the content seen at \( (u,v) \) in view \( j \) should appear in view \( i \). The value \( \alpha_{j \rightarrow i}(u,v) \) measures how reliable that match is. We discard correspondences whose confidence falls below a threshold. The result is a dense set of pixel-to-pixel matches across views that remains stable under wide viewpoint change.

\paragraph{Gaussian Initialization.}
We use the high-confidence correspondences to seed 3D Gaussian primitives directly in space. For a confident match between the pixel \( (u_j, v_j) \) in view \( j \) and its mapped location \( (u_i, v_i) \) in view \( i \), we back-project both pixels into 3D using known camera poses and triangulate their intersection. The resulting 3D point becomes the initial center of a Gaussian (see Figure~\ref{fig:pre-registration}, bottom right). Its initial appearance attributes are taken from the supporting image evidence, and its initial scale and orientation are set to cover a small spatial neighborhood around that 3D point. Repeating this over correspondences yields the initial Gaussian set \(\mathcal{G}_0=\{g_n\}\), 
where each \(g_n\) is a Gaussian primitive with position, scale, orientation, opacity, and color. Because these Gaussians are instantiated from dense correspondences rather than grown through iterative densification, \( \mathcal{G}_0 \) already provides broad and near-uniform spatial coverage of the scene. Subsequent geometric refinement adjusts these primitives but does not need to create a large number of new Gaussians.

\paragraph{Cross-view Context Association.}
The same correspondence field lets us record which masks from different views refer to the same scene content. Consider two masks \( M_i^a \) from view \( i \) and \( M_j^b \) from view \( j \). We project \( M_j^b \) into view \( i \) using the warp field \( W_{j \rightarrow i} \), producing a warped support mask in the coordinates of \( I_i \). We then measure how well this warped support overlaps \( M_i^a \), restricted to pixels with high correspondence confidence \( \alpha_{j \rightarrow i} \). If the overlap exceeds a threshold, we register a link that these two masks are consistent observations of the same underlying scene content. Repeating this procedure over view pairs accumulates the link set \(\mathcal{L}=\{(M_i^a,\tilde M_{j\to i}^b)\}\), 
where each pair in \(\mathcal{L}\) indicates strong cross-view agreement between two masks (see Figure~\ref{fig:pre-registration}, top right).

The pre-registration stage produces two artifacts. The first is an initialized Gaussian scene \( \mathcal{G}_0 \) created by triangulating dense correspondences. The second is a pool of mask links across views \( \mathcal{L} \) that captures which regions per-view act as the same scene content between viewpoints. Section~\ref{subsec:clustering} addresses how we cluster masks in \( \mathcal{L} \) into 3D Context Proposals.

\begin{figure*}[t]
    \centering
    \includegraphics[width=1.0\linewidth]{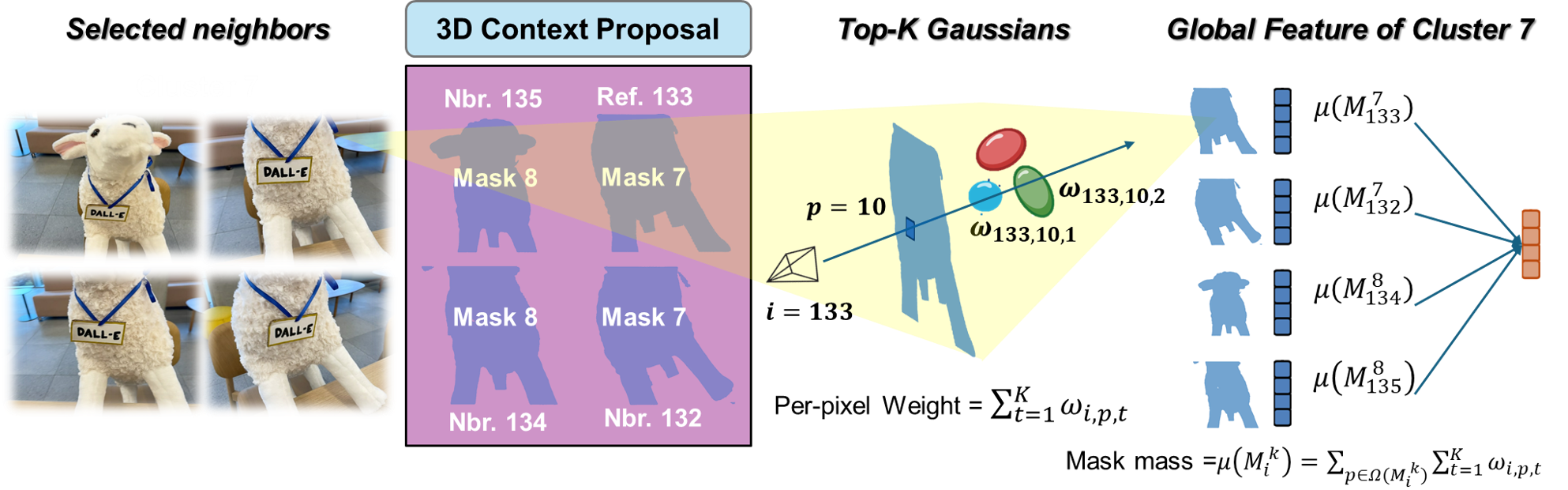}%
    \vspace{-2mm}
    \caption{\textbf{From context proposal to global feature.} Left: masks of the same entity are grouped into a 3D Context Proposal. Center: for a pixel $p$, the renderer returns the top-$K$ Gaussians with contributions $\{\omega_{i,p,t}\}_{t=1}^{K}$, from which the \emph{mask mass} $\mu\!\left(M_i^k\right)$ is computed. Right: a mass-weighted pool of member mask embeddings forms the proposal feature, which is registered to Gaussians via Eq.~(\ref{eq:accumulate}).}
    \label{fig:proposals}
    \vspace{-3mm}
\end{figure*}

\begin{algorithm}[t]
\caption{Cross-view mask clustering}
\label{alg:cluster}
\begin{algorithmic}[1]
\State \textbf{Inputs:} per-view sets $\{\mathcal{S}_i\}$ with $\mathcal{S}_i=\{(M_i^k, f_i^k)\}$; dense warp field  $W_{j\rightarrow i}$ and certainties $\alpha_{j\rightarrow i}$; visibility mask; thresholds $\tau_\alpha,\ \tau_{\text{iou}},\ \tau_{\text{box}}$; size gates $s_{\min},\ v_{\min}$.
\State Initialize graph $G=(V,E)$ with $V \gets \{(i,k)\ \forall\, M_i^k\}$, $E \gets \emptyset$
\ForAll{ordered view pairs $(i,j)$}
  \State $\Gamma_{j\rightarrow i} \gets [\alpha_{j\rightarrow i} \ge \tau_\alpha]\ \wedge\ \text{vis\_mask}$
  \ForAll{mask pairs $(M_i^a, M_j^b)$}
    \State $\widetilde{M}^{\,b}_{j\rightarrow i} \gets \mathcal{W}(M_j^b; W_{j\rightarrow i})$
    \State $O_{i,a;\,j,b} \gets \operatorname{IoU}( M_i^a \odot \Gamma_{j\rightarrow i},\ \widetilde{M}^{\,b}_{j\rightarrow i} \odot \Gamma_{j\rightarrow i} )$
    \State $\widetilde{M}^{\,a}_{i\rightarrow j} \gets \mathcal{W}(M_i^a; W_{i\rightarrow j})$
    \State $O_{j,b;\,i,a} \gets \operatorname{IoU}( M_j^b \odot \Gamma_{i\rightarrow j},\ \widetilde{M}^{\,a}_{i\rightarrow j} \odot \Gamma_{i\rightarrow j} )$
    \State $B_{i,a;\,j,b} \gets \operatorname{BBoxIoU}( M_i^a,\ \widetilde{M}^{\,b}_{j\rightarrow i} )$
    \State $B_{j,b;\,i,a} \gets \operatorname{BBoxIoU}( M_j^b,\ \widetilde{M}^{\,a}_{i\rightarrow j} )$
    \If{$O_{i,a;\,j,b}\ge\tau_{\text{iou}}$ \textbf{and} $O_{j,b;\,i,a}\ge\tau_{\text{iou}}$ \textbf{and} $B_{i,a;\,j,b}\ge\tau_{\text{box}}$ \textbf{and} $B_{j,b;\,i,a}\ge\tau_{\text{box}}$}
      \State Add undirected edge between $(i,a)$ and $(j,b)$ to $E$
    \EndIf
  \EndFor
\EndFor
\State Extract connected components $\{\mathcal{C}_m\}$ of $G$
\State Filter $\mathcal{C}_m$ by $|\mathcal{C}_m|\ge s_{\min}$ and $|\text{views}(\mathcal{C}_m)|\ge v_{\min}$
\State $\mathcal{P} \gets \{ P_m \equiv \mathcal{C}_m \}$
\State\Return $\mathcal{P}$
\end{algorithmic}
\end{algorithm}

\subsection{3D Context Proposals}
\label{subsec:clustering}

3D Context Proposals are formed through grouping per-view masks that mutually support one another under dense correspondence into stable multi-view units. 
We realize this by testing pairwise agreements under correspondence warps and linking masks that pass mutual gates; connected components in the resulting graph define the proposals. 

\paragraph{Cross-view Mask Clustering.}
Algorithm~\ref{alg:cluster} demonstrates the clustering procedure. Let a mask node be \(m=(i,k)\) with \(M_i^k \in \{0,1\}^{H\times W}\).
Given a candidate pair \((i,a)\) and \((j,b)\) with a dense warp \(W_{j\to i}\) from view \(j\) to \(i\) and a certainty map \(\alpha_{j\to i}\), we gate matches using a fixed certainty threshold \( \tau_\alpha \in [0,1] \) together with a renderer-derived visibility mask \texttt{vis\_mask}. The binary gate is defined as
\begin{equation}
\label{eq:gate}
\Gamma_{j\to i} \;=\; [\,\alpha_{j\to i} \ge \tau_\alpha\,] \;\wedge\; \text{vis\_mask}.
\end{equation}
The warped support in view \(i\) is obtained as
\begin{equation}
\label{eq:warp}
\widetilde{M}^{\,b}_{j\to i} \;=\; \mathcal{W}\!\big(M_j^b;\, W_{j\to i}\big),
\end{equation}
where \(\mathcal{W}\) denotes bilinear sampling at sub-pixel accuracy.
The confidence-gated overlap in view \(i\) is
\begin{equation}
\label{eq:overlap}
O_{i,a;\,j,b} \;=\; \operatorname{IoU}\!\Big(M_i^a \odot \Gamma_{j\to i},\; \widetilde{M}^{\,b}_{j\to i} \odot \Gamma_{j\to i}\Big).
\end{equation} We compute a coarse bounding-box agreement \(B_{i,a;\,j,b} = \operatorname{IoU}\!\big(\operatorname{box}(M_i^a),\, \operatorname{box}(\widetilde{M}^{\,b}_{j\to i})\big)\) and gate links with two thresholds, \(\tau_{\text{iou}}\) for mask overlap and \(\tau_{\text{box}}\) for box overlap. Agreement is required in both directions, and an undirected link is accepted only if
\begin{equation}
\label{eq:mutual-criteria}
\begin{aligned}
O_{i,a;\,j,b} &\ge \tau_{\text{iou}} \quad\text{and}\quad
O_{j,b;\,i,a} \ge \tau_{\text{iou}},\\
B_{i,a;\,j,b} &\ge \tau_{\text{box}} \quad\text{and}\quad
B_{j,b;\,i,a} \ge \tau_{\text{box}}.
\end{aligned}
\end{equation}
A graph \( G=(V,E) \) is then constructed with vertices \( V=\{(i,k)\} \). For every cross-view pair that passes the mutual gates above, we add an undirected edge to \( E \). The connected components of \( G \) define the raw proposals. Very small components are removed using two criteria: minimal member count \( s_{\min} \) and minimal distinct-view support \( v_{\min} \).
Each proposal \( P_m \) is represented only by its membership list \( (i,k) \), contributing view set, and compact per-view label maps for efficient lookup.

\subsection{Feature Registration}
\label{subsec:registration}

The goal of the registration stage is to assign a unit-normalized language descriptor to every Gaussian, enabling text queries to be evaluated directly in 3D. This stage operates on the initialized Gaussian set \(\mathcal{G}_0\), calibrated cameras, the per-view mask dictionary \(\mathcal{S}_i=\{(M_i^k,f_i^k)\}\), and the proposal set \(\mathcal{P}=\{P_m\}\) constructed in \S\ref{subsec:clustering}.

For a view \(i\) and a pixel \(p\), the renderer returns the indices and weights of the top-\(K\) Gaussians along the ray, denoted \(\{(g_{i,p,t},\,\omega_{i,p,t})\}_{t=1}^{K}\).
Their blending contributions are
\begin{equation}
\label{eq:ray-weights}
\begin{aligned}
\omega_{i,p,t} &= T_{i,p,t}\,\alpha_{i,p,t},\\[-2pt]
T_{i,p,t} &= \prod_{s< t}\bigl(1-\alpha_{i,p,s}\bigr),
\end{aligned}
\end{equation}
where \(\alpha_{i,p,t}\) is the effective opacity and \(T_{i,p,t}\) is the transmittance of the preceding Gaussians on the ray.

Each proposal \(P_m\) contains member masks drawn from multiple views. We compute a scalar \emph{mass} for every mask by integrating renderer contributions over the mask pixels
\begin{equation}
\label{eq:mask-mass}
\mu(M_i^k)=\sum_{p\in\Omega(M_i^k)}\ \sum_{t=1}^{K}\omega_{i,p,t}.
\end{equation}
The proposal descriptor is a mass-weighted pool of mask embeddings followed by \(\ell_2\) normalization,
\begin{equation}
\label{eq:proposal-desc}
\bar f_m=\frac{\sum_{(i,k)\in P_m}\mu(M_i^k)\,f_i^k}
               {\left\|\sum_{(i,k)\in P_m}\mu(M_i^k)\,f_i^k\right\|_2}.
\end{equation} An illustration of this aggregation is provided in Figure~\ref{fig:proposals}.

A pixel-wise proposal map \(L_i(p)\) is constructed for every training view, assigning each pixel inside a mask to the ID of its corresponding proposal in \(\mathcal{P}\). Pixels outside all masks receive a null label and are ignored. For each Gaussian \(g\in\mathcal{G}_0\), a feature accumulator \(A[g]\in\mathbb{R}^{D}\) and a scalar weight sum \(S[g]\in\mathbb{R}_{\ge0}\) are initialized to zero. For every pixel \(p\) with valid proposal \(m=L_i(p)\) and each of its top-\(K\) hits, the accumulation step is 
\begin{equation}
\label{eq:accumulate}
\begin{aligned}
A[g_{i,p,t}] &\gets A[g_{i,p,t}] + \omega_{i,p,t}\,\bar f_{m},\\[-2pt]
S[g_{i,p,t}] &\gets S[g_{i,p,t}] + \omega_{i,p,t}.
\end{aligned}
\end{equation} This registration step consumes the proposal feature from Figure~\ref{fig:proposals} and weights it by contributions $\omega_{i,p,t}$.

After processing all views, the descriptor for Gaussian
\(g\) is computed as
\begin{equation}
\label{eq:gaussian-feature}
f_g=\frac{A[g]}{\max(S[g],\varepsilon)}, 
\qquad
\hat f_g=\frac{f_g}{\|f_g\|_2},
\end{equation}
with a small \(\varepsilon\) for numerical stability. The implementation uses batched gather–scatter operations and relies only on renderer outputs.

\subsection{Inference Procedure}
\label{subsec:inference}
A text query is encoded to $f_q \in \mathbb{R}^D$ and normalized as $\hat f_q = f_q / \lVert f_q \rVert_2$.
Each Gaussian $g$ stores a registered descriptor from \S\ref{subsec:registration}. Following Dr.\ Splat \cite{drsplat25}, Product Quantization (PQ) is used for memory-efficient retrieval. Descriptors are stored as FAISS product-quantized codes and decoded to unit-normalized vectors at query time.

Cosine similarity is used to score Gaussians, \(\ s_g = \hat f_q^\top \hat f_g\). A FAISS PQ index over $\{\hat f_g\}$ produces a shortlist that is re-scored using decoded (full-precision) descriptors. Selection is performed directly in 3D without any render-based fine-tuning: a Gaussian is considered active if $s_g \ge \tau_{\text{act}}$. For visualization in view $i$, let $\{(g_{i,p,t},\,\omega_{i,p,t})\}_{t=1}^{K}$ denote the Top-$K$ contributors to pixel $p$. The activation mask is defined as
\begin{equation}
\label{eq:actmask}
M_i(p)=\mathbbm{1}[A_i(p)\ge\gamma], 
\end{equation}
where \(A_i(p)\) is the
sum of contributions over Top-\(K\) hits.

\begin{table*}[t]
\centering
\small
\setlength{\tabcolsep}{6pt}
\caption{Evaluation of 3D object selection on LERF-OVS \cite{lerf2023} dataset. Scores are averaged per scene and then across scenes. Bold indicates the best performance.}
\begin{tabular}{lccccc@{\hspace{10pt}}ccccc}
\toprule
& \multicolumn{5}{c}{mIoU $\uparrow$} & \multicolumn{5}{c}{mAcc@0.25 $\uparrow$} \\
\cmidrule(lr){2-6}\cmidrule(lr){7-11}
Method & waldo kitchen & figurines & ramen & teatime & mean
       & waldo kitchen & figurines & ramen & teatime & mean \\
\midrule
LangSplat     &  9.18 & 10.16 &  7.92 & 11.38 &  9.66  &  9.09 & 11.27 &  8.93 & 20.34 &  12.41 \\
LEGaussians   & 11.78 & 17.99 & 15.79 & 19.27 & 16.21  & 18.18 & 23.21 & 26.76 & 27.12 & 23.82 \\
OpenGaussian  & 24.57 & 53.01 & 24.44 & 55.40 & 39.36  & 36.36 & 83.93 & 39.44 & 76.27 & 59.00 \\
Dr.\ Splat    & 29.37 & 51.73 & 26.32 & 55.53 & 40.74  & 50.00 & 82.14 & \textbf{40.85} & \textbf{79.66} & 63.16 \\
\textbf{ProFuse (Ours)} & \textbf{36.91} & \textbf{56.13} & \textbf{28.16} & \textbf{62.78} & \textbf{46.00}
              & \textbf{68.18} & \textbf{85.71} & {39.44} & \textbf{79.66} & \textbf{68.25} \\
\bottomrule
\end{tabular}
\label{tab:objsel_miou_macc}
\end{table*}

\section{Experiments}
\begin{figure*}[t]
    \centering
    \includegraphics[width=1\linewidth]{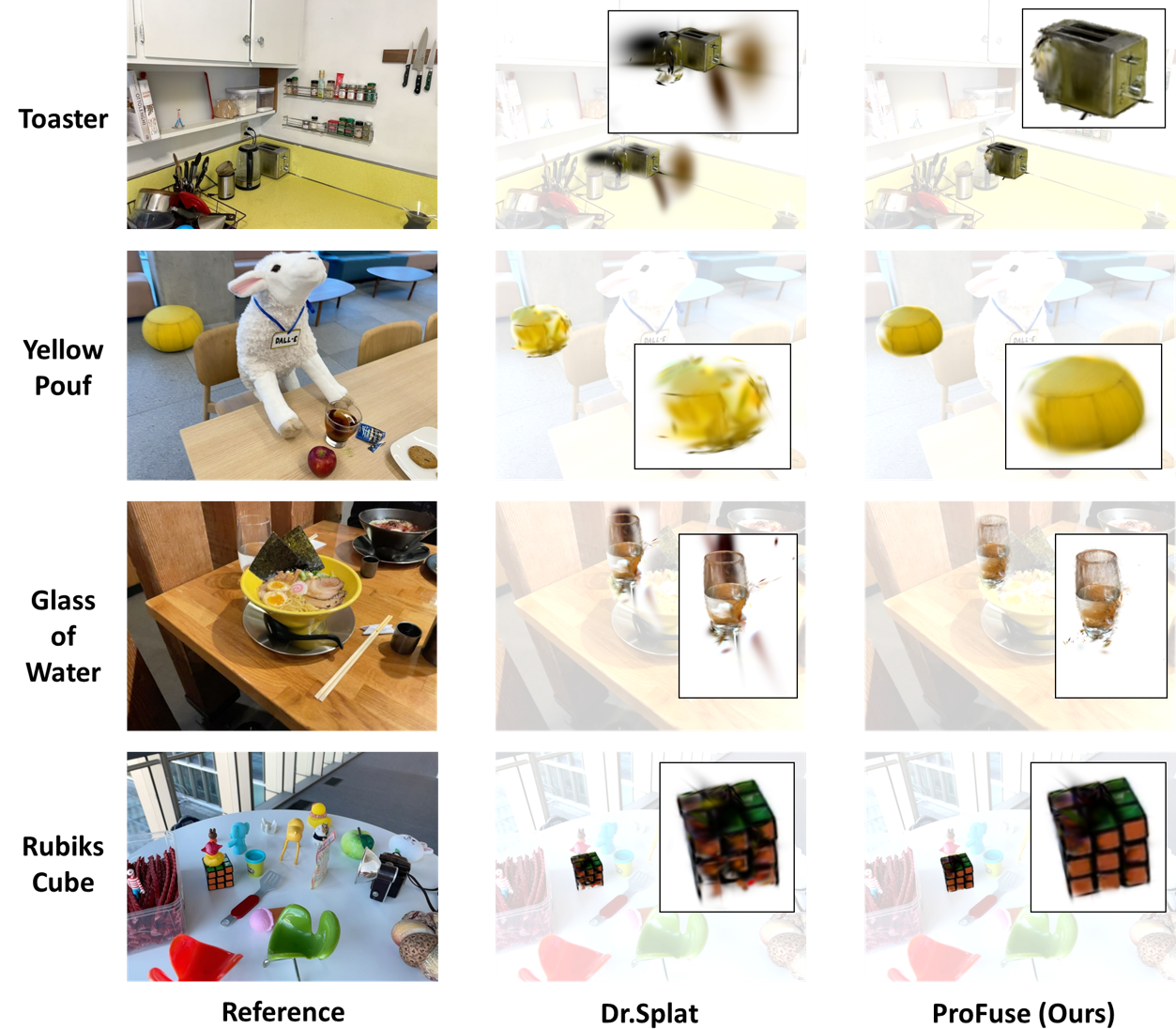}%
    \vspace{0mm}
    \caption{Qualitative comparison of object-level semantic queries on the LERF-OVS \cite{lerf2023} dataset. Our method produces more accurate and cleaner object retrieval, showing sharper correspondence between the text query and the selected 3D content.}
    \label{fig:qual_lerf}
    \vspace{1mm}
\end{figure*}

\subsection{Implementation}
Experiments are conducted on the LERF-OVS \cite{lerf2023} and ScanNet \cite{dai2017scannetrichlyannotated3dreconstructions} datasets. All four LERF scenes are used, and 10 scenes are sampled from the ScanNet dataset. SAM-based segmentation and mask embedding are preprocessed on 8 NVIDIA H100 GPUs, while all remaining experiments run on a single A100 GPU.

\begin{figure*}[t]
    \centering
    \includegraphics[width=\linewidth]{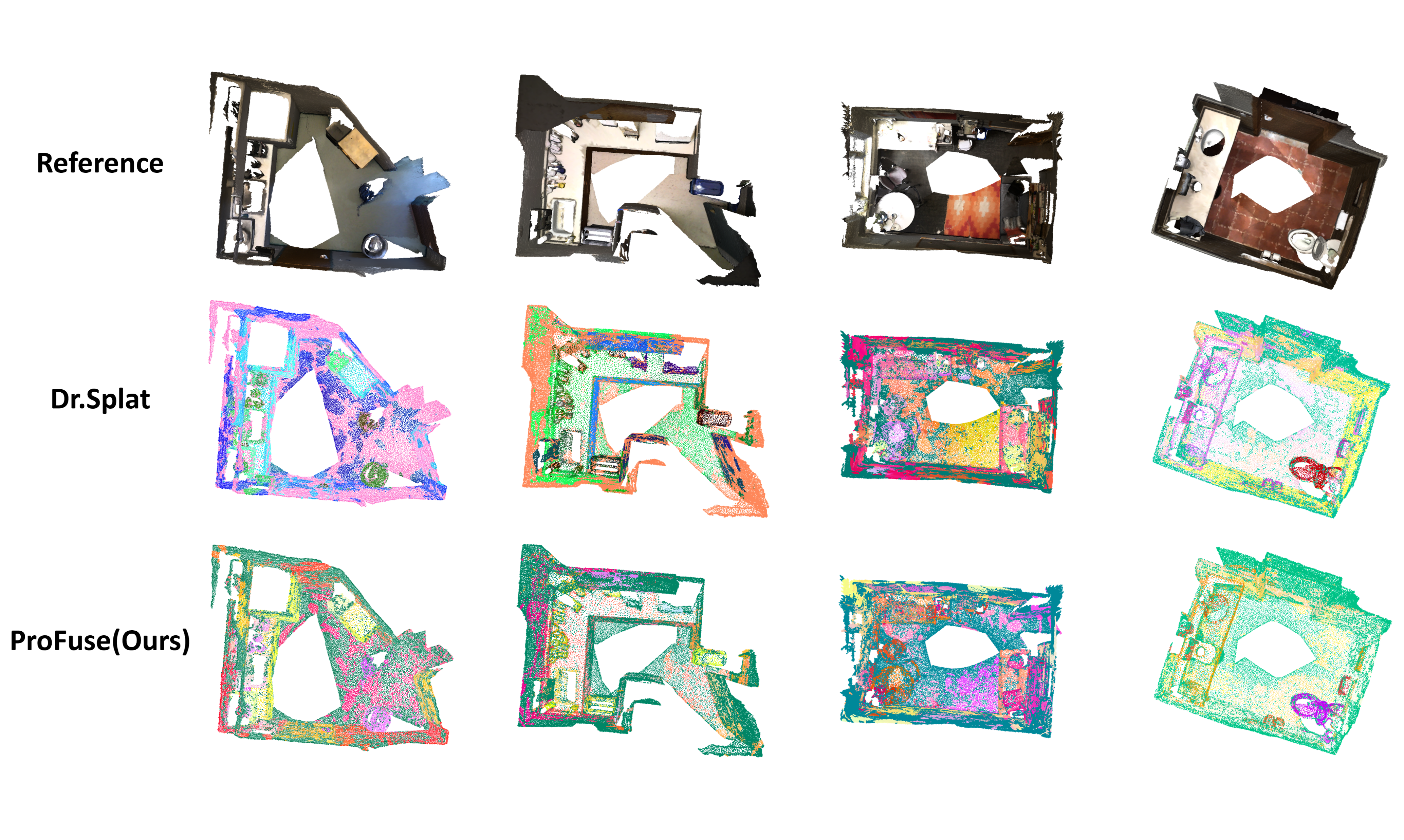}%
    \vspace{-3mm}
    \caption{Feature visualizations on the ScanNet \cite{dai2017scannetrichlyannotated3dreconstructions} dataset using registration-based methods. Colors represent normalized language features transferred to mesh vertices and rendered via a fixed RGB projection. ProFuse produces cleaner regions with sharper boundaries and fewer speckles.}
    \label{fig:qual_scannet}
    \vspace{-3mm}
    \label{sec:objsel}
\end{figure*}

\subsection{Open-Vocabulary 3D Object Selection}

We evaluate open-vocabulary 3D object selection on the four LERF scenes using the official text queries and splits. Each method outputs a binary activation per frame, while our pipeline performs selection directly in 3D.
Let $q\!\in\!\mathbb{R}^D$ be the CLIP text embedding, normalized as $\hat q\!=\!q/\lVert q\rVert_2$. Each Gaussian \( g \) stores a normalized language feature $\hat f_g$ from registration. Active Gaussians are defined as \(\mathcal{G}_\tau=\{\, g \;|\; \langle \hat f_g,\hat q\rangle\ge \tau \,\},\) with a method-specific global threshold $\tau$. For view $i$ and pixel $p$, the renderer provides the top-$K$ Gaussians and weights $\omega_{i,p,t}$ . The activation is
\[
A_i(p)=\sum_{t=1}^{K}\,\omega_{i,p,t}\,\mathbf{1}\!\left[g_{i,p,t}\in \mathcal{G}_\tau\right],
\]
and the mask is $\widehat{M}_i=\mathbf{1}[\,A_i\ge \gamma\,]$ using a fixed silhouette threshold $\gamma$. A small grid search is used to determine the global threshold $\tau$ for each method. \emph{mean IoU} is computed by evaluating intersection-over-union for each query–frame pair and averaging across all queries and frames in a scene. The final score is obtained by averaging across the four scenes. Table~\ref{tab:objsel_miou_macc} reports these quantitative results. The metric \emph{mAcc@0.25} is also provided, defined as the fraction of query–frame pairs with IoU at least $0.25$, using the same $\tau$.

Qualitative results are presented in Figure~\ref{fig:qual_lerf}. Our method isolates the queried object with far fewer background activations, yielding cleaner and more semantically precise selections. In contrast, Dr.\ Splat often exhibit ray-like spillovers into nearby clutter or textured areas. For instance, the “Toaster” query incorrectly highlights the entire kettle on the left, while the “Glass of Water” query becomes distracted by specular reflections.

\begin{table}[t]
\centering
\caption{Open-vocabulary point cloud understanding on ScanNet. Results use mIoU and mAcc for 19/15/10-class settings.}
\label{tab:points_pc}
\resizebox{\columnwidth}{!}{%
\begin{tabular}{lcc|cc|cc}
\toprule
\multirow{2}{*}{Method} &
\multicolumn{2}{c|}{19 classes} &
\multicolumn{2}{c|}{15 classes} &
\multicolumn{2}{c}{10 classes} \\
& mIoU$\uparrow$ & mAcc$\uparrow$ & mIoU$\uparrow$ & mAcc$\uparrow$ & mIoU$\uparrow$ & mAcc$\uparrow$ \\
\midrule
LangSplat        &  3.78 &  9.11 &  5.35 & 13.20 &  8.40 & 22.06 \\
LEGaussians      &  3.84 &  10.87 &  9.01 & 22.22 &  12.82 & 28.62 \\
OpenGaussian     & 24.73 & 41.54 & 30.13 & 48.25 & 38.29 & 55.19 \\
Dr.\ Splat       & 28.40 & 52.77 & 32.67 & 58.53 & 36.81 & 66.41 \\
\textbf{ProFuse (Ours)} & \textbf{30.52} & \textbf{55.32} & \textbf{34.76} & \textbf{60.90} & \textbf{39.74} & \textbf{69.38} \\
\bottomrule
\end{tabular}%
}
\label{tab:table2}
\end{table}

\subsection{Open-Vocabulary Point Cloud Understanding}
The evaluation is conducted on the ScanNet dataset using the label spaces defined in OpenGaussian \cite{wu2024opengaussian}, considering class sets of 19, 15, and 10 categories. Each mesh vertex in the aligned reconstruction is assigned a semantic label, and class names are encoded once into language embeddings and reused across all methods.

Per-Gaussian language codes are first decoded using FAISS PQ to obtain cosine logits against class embeddings. These logits are transferred to mesh vertices through a spatially aware kernel that respects each Gaussian’s full ellipsoid. Candidate Gaussians are shortlisted by Euclidean proximity ($K{=}64$), filtered by an elliptical Mahalanobis gate ($\sigma{=}3$), and weighted by both $\exp(-\tfrac{1}{2}d^2)$ and Gaussian opacity. A \emph{softmax} over class logits yields per-candidate class probabilities, and vertex scores are computed as the weighted sum of all candidates. Because predictions occur directly in 3D, no rendering is involved during evaluation. The same kernel and shortlist configuration is applied to every method so that performance differences reflect the quality of the learned Gaussian features rather than variations in the transfer rule. Ten scenes from ScanNet are sampled for evaluation, and scores are computed with fixed hyperparameters to report average \emph{mIoU} and \emph{mAcc} for each class set. Quantitative results for the 19-, 15-, and 10-class settings are provided in Table ~\ref{tab:table2}. 

To contextualize point-level scores, we visualize feature colorings of ScanNet reconstructions and compare them to the pioneer registration-based baseline Dr.\ Splat \cite{drsplat25} in Figure~\ref{fig:qual_scannet}. For each scene, we show the reference mesh view and two pseudo-colored point clouds. Colors are obtained by projecting normalized per-Gaussian features to three channels and painting the transferred per-vertex features; views are matched to the reference for consistent framing. Dr.Splat tends to produce darker, patchy fragments and color bleeding near corners, whereas our results exhibit higher region consistency with large surfaces rendered in coherent color swaths. We achieve cleaner boundaries at furniture edges and fixtures with fewer mixed colors at object–wall contacts.

\subsection{Training Efficiency}

\begin{table}[t]
\centering
\footnotesize
\caption{Comparison of training requirements and retrieval speed across 3D scene understanding methods.}
\setlength{\tabcolsep}{3.8pt}
\renewcommand{\arraystretch}{0.95}
\resizebox{\columnwidth}{!}{\begin{tabular}{lcccc}
\toprule
Method & Scene & Render supervision & Feature distill. & Query \\
\midrule
LERF           & NeRF          & required & $\sim$24 h & slow \\
LangSplat      & SfM--3DGS     & required & $\sim$4 h  & slow \\
LEGaussians    & SfM--3DGS     & required & $\sim$4 h  & slow \\
OpenGaussian   & SfM--3DGS     & required & $\sim$1 h  & fast \\
GOI            & SfM--3DGS     & required & $\sim$12 min & fast \\
Dr.\ Splat     & SfM--3DGS     & none & $\sim$10 min & fast \\
\textbf{ProFuse (Ours)} & \textbf{Corr-init 3DGS} & \textbf{none} & \textbf{$\sim$5 min} & \textbf{fast} \\
\bottomrule
\end{tabular}}
\label{tab:discrete-efficiency}
\vspace{-2mm}
\end{table}

\begin{table}[t]
\centering
\footnotesize
\caption{Wall-clock comparison of geometry, semantic processing, and indexing time on the LERF dataset.}
\setlength{\tabcolsep}{4pt}
\renewcommand{\arraystretch}{0.95}
\resizebox{\columnwidth}{!}{
\begin{tabular}{lccccc}
\toprule
Method & Geometry & Semantics & Total & Indexing\\
\midrule
OpenGaussian     & $\sim$20 m  & $\sim$ 40 m          & $\sim$1 h & Codebook\\
Dr.\ Splat       & $\sim$20 m  & $\sim$ 0 + 10 m      & $\sim$30 m & PQ\\
\textbf{ProFuse (Ours)} & \textbf{$\sim$2 + 15 m} & \textbf{$\sim$2m + 20 s} & \textbf{$\sim$19 m} & PQ\\
\bottomrule
\end{tabular}}
\label{tab:time_breakdown}
\vspace{-2mm}
\end{table}

\begin{table}[t]
\centering
\caption{Top-$K$ analysis on ScanNet showing mIoU and feature registration time for registration-based methods.}
\resizebox{\columnwidth}{!}{%
\begin{tabular}{lcc|cc|cc}
\toprule
\multirow{2}{*}{Method} &
\multicolumn{2}{c|}{Top $K{=}10$} &
\multicolumn{2}{c|}{Top $K{=}20$} &
\multicolumn{2}{c}{Top $K{=}40$} \\
& mIoU$\uparrow$ & time$\downarrow$ & mIoU$\uparrow$ & time$\downarrow$ & mIoU$\uparrow$ & time$\downarrow$ \\
\midrule
Dr.\ Splat       & 33.82 & $\sim$45 s & 35.57 & $\sim$85 s & 36.81 & $\sim$165 s \\
\textbf{ProFuse (Ours)} & \textbf{39.74} & \textbf{$\sim$25 s} & \textbf{39.74} & \textbf{$\sim$25 s} & \textbf{39.74} & \textbf{$\sim$25 s} \\
\bottomrule
\end{tabular}%
\label{tab:topk_miou}
}
\end{table}

The cost of attaching open-vocabulary semantics to a reconstructed scene is measured in wall-clock time. As shown in Table~\ref{tab:discrete-efficiency}, render-supervised distillation methods require hours of processing, and existing registration-based approaches \cite{drsplat25} still take several minutes. ProFuse achieves the fastest runtime through correspondence-guided initialization, which produces a compact Gaussian set without densification, and through lightweight proposal-level feature fusion. These components reduce semantic attachment to about five minutes per scene, making ProFuse 2× faster than the prior SOTA. Table~\ref{tab:time_breakdown} provides a runtime breakdown of direct 3D methods. ProFuse reduces scene-specific semantic association to only a few minutes because proposal construction is lightweight and registration uses simple contribution accumulation without gradient updates. The compact geometry from correspondence-guided initialization removes densification and further shortens processing time.

\subsection{Ablation Study}
To isolate the effect of correspondence-guided geometry and context proposals, we study the impact of the Top-K Gaussian candidates used during feature registration. Table~\ref{tab:topk_miou} reports mIoU and registration time on ScanNet under three settings \(K{=}10, 20, 40\). Without context proposals, registration-based baselines typically require \(K{=}40\) to achieve saturation, indicating weak concentration of semantic mass along the viewing ray. In contrast, ProFuse reaches its maximum accuracy with \(K{=}10\). The global proposal features place most of the mass on the leading few Gaussians, while our correspondence-initialized geometry further reduces long-tail ambiguity. As a consequence, larger \(K\) offers no additional benefit, and a compact \(K{=}10\) is sufficient for both accuracy and speed.

\section{Conclusion}
ProFuse enforces cross-view semantic consistency in 3DGS without requiring any render-supervised learning for semantics. Dense correspondences generate 3D Context Proposals, and visibility-weighted fusion yields a coherent semantic field. Experiments on LERF and ScanNet confirm accurate open-vocabulary selection and point-level understanding, showing that correspondence-guided geometry provides an efficient path to semantic association in 3DGS.

{
    \small
    \bibliographystyle{ieeenat_fullname}
    \bibliography{main}
}


\clearpage
\appendix


\section*{Supplementary material}

# ProFuse: Efficient Cross-View Context Fusion for Open-Vocabulary 3D Gaussian Splatting

Supplementary Material

## A. Reproducibility and Code Release

The implementation of ProFuse will be released at https://github.com/chiou1203/ProFuse. This repository will include the training code for the dense correspondence guided Gaussian initialization, cross-view mask clustering, and feature registration stages.

## B. Discussion

### B.1. Limitation

Although cross-view mask clustering helps associate masks that likely refer to the same context, our method still remains bounded by SAM and CLIP. Mask embeddings still reflect the underlying segmentation quality. Moreover, even with accurately segmented masks, CLIP embedding errors directly affect 3D scene understanding, particularly for similar objects and uncommon text prompts.

Though pre-registration is correspondence driven, even with warps, mismatches can persist when masks are imperfect. For example, under an IoU threshold of 0.5, a mask that is 80% a large object and 20% a small object may be grouped with other masks of the large object; the pooled global feature then inherits contamination from the small object. This relates to over-coarse grouping in which masks group that are not clean and highly accurate washed out fine details and an inaccurate global feature can pollute all its mask members.

### B.2. Societal Impact

The method lowers the barrier to open vocabulary understanding in 3D scenes. It attaches language descriptors to Gaussians through correspondence driven pre-registration and feature registration without a render supervised loop. The result is shorter iteration time and a lighter compute footprint. These gains translate into practical uses. Education and cultural heritage benefit from interactive exploration of reconstructed spaces where a user can ask for an object and see it in context. AR and VR authors gain a searchable index over large captures that enables precise selection and editing without project specific training. Robotics and digital twins obtain faster scene lookup for inventory, maintenance, and task setup in indoor environments. Assistive scenarios become more responsive since a user can request a target item and receive immediate guidance in a captured room.

Responsible deployment remains straightforward. Capture and indexing should follow clear consent. Storage and sharing should use established governance in each setting. With these norms in place, ProFuse helps democratize semantic interaction with 3D content and broadens access to practical tools for learning, creation, and operation.

## C. Preliminaries

### C.1. 3D Gaussian Splatting

3D Gaussian Splatting represents a scene with a set of anisotropic Gaussians. Each primitive has a mean in world space and a covariance that is factorized into a rotation and a diagonal scale. This factorization guarantees a valid positive semi-definite matrix and is convenient for optimization.

Rendering proceeds by projecting each 3D covariance to image space through a first-order camera Jacobian, which yields a 2×2 covariance for splatting on the raster plane. The pixel color is then obtained by front-to-back alpha compositing. The formulation matches volumetric rendering and can be written as a sum of per-splat contributions, where the contribution of the i-th splat equals its transmittance times its effective opacity times its color. Transmittance accumulates along the ray as the product of one minus the previous opacities.

The original system uses a differentiable tile-based rasterizer. Gaussians are culled against the frustum and tiles, sorted by depth, and blended per tile to maximize parallelism while maintaining the same alpha-compositing model.

### C.2. Product Quantization (PQ)

ProFuse follow Dr.Splat who utilize Product Quantization to store and search language features efficiently without per-scene codebook training. PQ partitions a D-dimensional vector into L sub-vectors, learns a codebook per subspace, and represents each sub-vector by the index of its nearest centroid. This reduces memory and turns distance or similarity computation into table lookups across subspaces.

After training centroids, a lookup table stores all pairwise distances among centroids in each subspace. The distance between two PQ-encoded vectors becomes a sum

of L table entries, one per subspace. Cosine similarity can be computed in the same way using inner-product tables after normalizing subvectors. This design shifts the cost from high-dimensional arithmetic to indexed retrieval while preserving correlation with true distances within known quantization bounds.

Significant search-time gains over direct cosine similarity on CLIP features can be observed when varying the sub-vector size. These measurements demonstrate that LUT-based PQ search scales well for large 3D Gaussian sets and supports interactive text-to-3D queries.

## D. Implementation Details

### D.1. Correspondence Driven Gaussians

Unlike standard 3DGS, we initialize the scene from dense cross-view correspondences and then perform a pruning-only optimization without densification. A trainer calls a correspondence-based initializer and proceeds with a photometric objective under a fixed training schedule.

We first sample a compact set of reference views by K-means clustering in pose space to cover the trajectory with minimal redundancy. For each reference we attach a small pose-nearest neighbor set, selected by distance in the same pose space. We sample a set of reference views and attach a small pose-nearest set of neighbors to each reference. In our runs, we use 180 reference views and 3 neighbors per reference for maximum efficiency. The initializer computes a dense warp field and certainties of each warp, aggregates the most confident warp per pixel, and triangulates 15,000 correspondences per reference to seed Gaussians. The seed carries position, color, and scale from the paired views, followed by standard splat optimization. We enable pruning while disabling densification during training, removing poor splats without ever growing new ones. Opacity resets are effectively off through a very large reset interval. We use a batch size of 64 and run 30,000 iterations per scene. This procedure typically yields around $2\times10^6$ initialized Gaussians and roughly $5\times10^5$ to $10^6$ active Gaussians after pruning

### D.2. The ProFuse Framework

ProFuse attaches language descriptors to a Gaussian scene that is initialized from dense correspondences and refined with pruning only. The representation follows standard 3D splatting for geometry and visibility. The semantic path operates on masks and text features and produces a per-Gaussian descriptor that supports open-vocabulary queries without a render-supervised loop.

Masks come from the object level of SAM. Each mask is encoded by CLIP ViT-H/14 with a 512-dimensional embedding. Per-view features are fused to Gaussians using the Top-K ray contributions from the renderer so the same weights that produce color also produce language features. The fusion creates a single descriptor per Gaussian that is shared across views and does not depend on any prompt at training time.

Descriptors are stored with Product Quantization. We use a global codebook that is shared across all scenes and keep the PQ codes as the only per-Gaussian semantic payload during training and inference. The system reconstructs codes to unit-norm vectors for cosine scoring when answering a text query. This design reduces memory and enables fast similarity evaluation while keeping the scoring rule identical to the one used for visualization and selection.

### D.3. Compare Model Settings

**LangSplat.** LangSplat learns a 3D language field on 3D Gaussians and replaces NeRF rendering with tile-based splatting for language features. It builds a scene-wise language autoencoder and trains language features in a scene-specific latent space rather than directly on CLIP space, which reduces memory. Supervision comes from SAM to form hierarchical semantics so that subpart, part, and whole concepts are separable. The paper reports large speed gains over LERF at high resolution, which is consistent with the splatting design and the latent-space training.

**LEGaussians.** LEGaussians discretizes language features with a learnable codebook and stores indices rather than full float descriptors. Quantization selects the nearest basis in a discrete feature space using a CLIP term and a controllable DINO term; optimization aligns dense image features to their quantized counterparts. During training the method renders compact semantic vectors from Gaussians and decodes them with a small MLP under a cross-entropy objective. It further adds adaptive spatial smoothing driven by a learned per-Gaussian uncertainty so that semantics vary smoothly where features are unstable. These design choices reduce storage and regularize multi-view inconsistency.

**OpenGaussian.** OpenGaussian augments each Gaussian with a low-dimensional instance feature and learns it by rendering feature maps with alpha blending. Supervision uses SAM boolean masks without cross-view correlation. The loss encourages intra-mask smoothness and inter-mask separation so that features within an object cluster together while different objects separate in feature space. To discretize for efficient retrieval, a two-level codebook is constructed in a coarse-to-fine manner. The coarse stage clusters by concatenating position with features; the fine stage clusters by features only, which preserves geometry and improves scalability in larger scenes. The paper also proposes an instance-level association that links 2D CLIP to 3D points without additional training.

**Dr. Splat.** Dr. Splat performs direct feature registration on pre-trained 3DGS scenes. Per-pixel CLIP embeddings are aggregated onto the dominant top-k Gaussians along each camera ray with weights equal to transmittance times effective opacity from the volume rendering equation. The aggregated embeddings are product-quantized and stored as PQ indices, enabling compact storage and fast 3D search without per-scene feature distillation. The paper contrasts this registration-based pipeline with rendering-supervised methods and reports substantially shorter end-to-end preparation and query times.

### D.4. Training Details

**Dense Correspondence.** The settings for dense correspondence are shown in Table 1. We select 180 reference views by K-means in pose space and attach 3 pose-nearest neighbors to each reference by k-NN. We use RoMa as the pretrained network for dense matches. We cap the sampling at 15,000 matches per reference. Certainty is aggregated by a per-pixel maximum across neighbors.

**Gaussian Initialization.** The Gaussian initialization and optimization settings are summarized in Table 2. Each correspondence track is triangulated with calibrated cameras, and we keep only tracks whose mean reprojection error is below 0.01 in normalized image coordinates. For every surviving 3D point we create a Gaussian with spherical covariance, where the initial scale parameters are set to 0.001 in scene units.

**Gaussian Optimization.** The initialized scene is then optimized with the 3DGS training loop for 30,000 iterations with batch size 64. The position learning rate starts at $1.6\times10^{-4}$ and decays to $1.6\times10^{-6}$ over 30,000 steps with a delay multiplier of 0.01. The feature learning rate is 0.0025, the opacity learning rate is 0.025, the scaling learning rate is 0.005, and the rotation learning rate is 0.001. We keep the dense ray sampling ratio at 1 % of pixels per iteration (percent_dense = 0.01) and weight the DSSIM term by 0.2. Densification itself is disabled by setting no_densify to True. The opacity reset interval is extended to 1,000,000 iterations so that no opacity reset occurs during optimization.

**Cross-View Mask Clustering.** After dense feature matching, we cluster masks that likely depict the same scene region across views. For each reference image, we use SAM mask level 1 as the object-level partition. The reference segmentation is resized to the RoMa canvas resolution, and each neighbor segmentation is projected into this canvas using the warp. We prune projected labels with the Gaussian visibility mask, using a transmittance threshold of 0.05. The warped IoU threshold for cluster edges are set to 0.2 and bounding box IoU 0.08. Very small masks that cover less than 0.5% of the canvas use a stricter IoU requirement of 0.30 in order to avoid spurious links caused by noise. Connected components in this graph define the cross-view clusters, and for each reference image we store the cluster assignments and corresponding SAM label ids in a NPZ file that is reused by the registration stage.

Table 1: Dense correspondence setting.

| Config | Value |
| --- | --- |
| Dense Matching Network | RoMa |
| Total References | 180 |
| Neighbors per Reference | 3 |
| Max Matches per Reference | 15,000 |

Table 2: Gaussian pre-training setting.

| Config | Value |
| --- | --- |
| Triangulation Reprojection Tolerance | 0.01 |
| Initial Scale | 0.001 |
| Training Iterations | 30,000 |
| Batch Size | 64 |
| Optimizer | Adam |
| Base Position Learning Rate | $1.6\times10^{-4}$ |
| Delay Multiplier | 0.01 |
| Feature Learning Rate | 0.0025 |
| Opacity Learning Rate | 0.0025 |
| Scaling Learning Rate | 0.005 |
| Rotation Learning Rate | 0.001 |
| Percent Dense Pixels | 0.01 |
| DSSIM weight | 0.2 |
| Densification | Disabled |

Table 3: Cross-view mask clustering setting.

| Config | Value |
| --- | --- |
| SAM Mask Level | 1 (Object) |
| Visibility Threshold | 0.05 |
| Warped IoU | 0.2 |
| Bounding Box IoU | 0.08 |
| Small Mask Fraction | 0.005 |
| Small Mask IoU | 0.3 |
| Edge Requirement | Mutual Best Neighbor |

Table 4: Feature registration setting.

| Config | Value |
| --- | --- |
| Feature Level | 1 |
| Top-K | 10 |
| PQ Index | 128-D |
| Pixel Stride | 1 |
| SpMM Cluster Block | 0 |
| Eps Contribution | 0 |

**Feature Registration.** Settings for registration are summarized in Table 4. During registration, we freeze the Gaussian scene and run a single registration pass that attaches a 512-dimensional language descriptor with feature level 1 to every Gaussian. Proposal-level feature registration is used whenever a valid NPZ file is available and the code falls back to per-mask accumulation only when a view has no mapped cluster metadata.

For each camera, we render the top-10 Gaussian ids and their contributions per pixel from the pretrained scene. To control sampling density on clustered pixels, we introduce pixel stride. Valid pixels inside clustered masks are sub-sampled with a uniform stride in image space, keeping the option for heavier scenes if needed. After collecting top-10 contributions for these pixels, we build a sparse weight matrix between Gaussians and cluster ids. Before forming this matrix, we apply a small threshold on contributions through eps_contrib. Entries with contribution below this threshold are discarded, which removes numerically tiny pairs that only add memory cost but almost no semantic signal. We set default stride to 1 and eps_contrib to 0 in reported experiments.

The accumulation is implemented as a sparse matrix–dense matrix multiplication over Gaussians and global features. An optional parameter spmm_cluster_block allows the cluster axis to be processed in blocks when GPU memory is tight. Each block builds a smaller sparse matrix for a subset of clusters and accumulates the result into the global Gaussian buffers. We set spmm_cluster_block = 0 in all our runs since our scenes fit comfortably within memory at this resolution.

Once the per-Gaussian float features are obtained, we normalize them and encode them with product quantization. We always enable PQ and load a pretrained FAISS index with code size 128. The final stored language descriptor of each Gaussian is a 128-byte PQ code.

### D.5. Evaluation Details

**Open-Vocabulary 3D Object Selection.** We adopt the LERF object selection benchmark and use the same four scenes, prompts, and binary ground-truth masks as in Section 4.2. For each method we compute cosine similarity between every Gaussian language descriptor and the CLIP text embedding of the query. Gaussians are activated when the similarity exceeds a threshold $\tau$. In practice, $\tau$ is chosen for each method by grid search, sweeping values in steps of 0.01 and fixing the best value across all scenes. The re-ranking stage follows the LERF relative-relevance formulation with the canonical word list {object, things, stuff, texture} and temperature $\tau_{\text{rerank}} = 3.0$. After selecting Gaussians, we aggregate their per-pixel contribution weights thresholding at a contribution level $\gamma = 0.025$. We then compute mIoU and mAcc@0.25 on the same set as LERF and average the scores over all scenes.

**Open-Vocabulary 3D Point Cloud Understanding.** For ScanNet dataset, we evaluate all methods on the 19-class, 15-class, and 10-class label sets with a strict point-level protocol. For each checkpoint we decode the 512-dimensional PQ language features back to float vectors using the original FAISS index. Codes that are all-255 or mapped to an invalid IVF list are treated as invalid, decoded features are L2-normalized, and invalid rows are kept as zeros. Class text features are loaded from an JSON, aligned to the class name list for each label set, and L2-

Table 5: Evaluation setting.

| Config | Value |
| --- | --- |
| Contribution Threshold | 0.025 |
| Reranking Temperature | 3 |
| Gaussian Candidates | 64 |
| Gating Radius | 3 |
| SoftMax Temperature | 1 |

normalized. We form cosine logits between all Gaussians and all class text features and transfer these logits to mesh vertices using the rotation- and scale-aware Mahalanobis kernel with opacity weighting. The kernel uses a shortlist of $k_{\text{shortlist}} = 64$ Gaussian candidates per point, a gating radius $\sigma_{\text{gate}} = 3.0$ in Mahalanobis distance, and a SoftMax temperature logit_temp = 1.0; these values are shared by the 19-class, 15-class, and 10-class evaluators. Points that fall outside the $\sigma$gate fall back to the nearest valid Gaussian. We report point-mIoU and point-mAcc averaged over classes that are present in the ground truth and over the evaluation scenes listed in Section 4.3.

### D.6. Computing Resource Configuration

The experiments of ProFuse were conducted on a single NVIDIA A100 80 GB GPU. All methods were evaluated with their best model on their best threshold to maintain consistency. ScanNet scenes are down sample to 170 ∼ 210 images per scene, and each Gaussian scene was trained for 30k iterations with the same hyper-parameter setting during scene optimization. The experiments on registration-based methods were compared using the same PQ codebook with sub-vector size 128, and best threshold were picked method-wise to respect different model nature, ensuring a comprehensive and uniform assessment of performance across different architectures.

# E. Additional Experiments

## E.1. Gaussian Scene Experiments

We investigate how the choice of pretrained Gaussian scene affects both reconstruction quality and semantic association. Three variants are evaluated under the same ProFuse registration pipeline and object–selection protocol, as summarized in Tables 6–9. The first variant applies ProFuse on top of a standard SfM-based 3D Gaussian Splatting scene. Dense correspondence and cross-view mask clustering are still computed, but these tracks only influence the semantic side; the underlying geometry follows the original 3DGS training procedure. The second variant is the default ProFuse configuration used in the main paper. In this case the Gaussian scene is initialized from dense correspondence tracks without densification, followed by our 30k-step pruning-only optimization. The third variant enables densification on top of the correspondence-guided seeds.

Scene-level mIoU and mAcc@0.25 for the LERF object-selection task are reported in Tables 6 and 7. The correspondence-guided scene without densification consistently improves over the pure 3DGS scene in both metrics, and achieves the best mean semantic performance across the four scenes. Enabling densification yields mixed behavior. Some scenes remain competitive, while others suffer from a noticeable drop in mIoU and mAcc. We attribute this to the strong variation of the similarity–threshold curve across scenes. A single global activation threshold is applied within each variant, which is a reasonable choice for comparison but cannot simultaneously track the per-scene optimum once the density pattern of Gaussians changes significantly.

Table 8 reports PSNR of the rendered views from the three pretrained scenes. Here the correspondence-guided initialization with densification achieves the highest mean PSNR, while the non-densified ProFuse scene is slightly below. This indicates that densification still brings benefits for pure reconstruction, even when the initial seeds already introduce highly accurate geometry.

Table 9 further compares the optimization time across scenes. The correspondence-guided scene without densification shortens training to roughly 14 minutes on average. Enabling densification increases the mean time to about 22 minutes, almost doubling the cost on some scenes. Combining these trends, the default ProFuse configuration, which uses correspondence-guided seeds without densification, forms a practical compromise. It delivers the strongest semantic performance in the object-selection benchmark, preserves competitive reconstruction quality, and keeps the pretraining time significantly lower than the densified alternative.

Qualitative example of the scene reconstruction progress for correspondence-guided 3DGS is illustrated in Figure 1. The left shows the reference image. The second column renders the raw seeds produced directly from dense correspondence before any optimization, already capturing the layout of major objects. The third column shows the scene after 7k optimization steps of pruning, where geometry and appearance become noticeably sharper. The right column viualize the result after 30k iterations, which mainly refines shading and small details. This qualitative behavior is consistent with the quantitative results and illustrates that correspondence-guided initialization provides a strong geometric prior even without densification

Table 6: Ablation study on scene mIoU for different pretrain Gaussian scenes.

| Gaussian Scene | Scene mIoU | | | | |
| --- | --- | --- | --- | --- | --- |
| | Waldo kitchen | Figurines | Ramen | Teatime | Mean |
| 3DGS | 24.45 | 55.27 | 24.70 | 62.57 | 41.75 |
| Corr-init | 36.91 | 56.13 | 28.16 | 62.78 | 46.00 |
| Corr-init + Densify | 14.29 | 41.31 | 28.16 | 60.62 | 36.10 |

Table 7: Ablation study on scene mAcc for different pretrain Gaussian scenes.

| Gaussian Scene | Scene mAcc@0.25 | | | | |
| --- | --- | --- | --- | --- | --- |
| | Waldo kitchen | Figurines | Ramen | Teatime | Mean |
| 3DGS | 36.36 | 83.93 | 40.85 | 79.66 | 59.85 |
| Corr-init | 68.18 | 85.71 | 39.44 | 79.66 | 68.25 |
| Corr-init + Densify | 22.72 | 67.86 | 39.44 | 79.66 | 52.42 |

Table 8: Ablation study on scene PSNR for different pretrain Gaussian scenes.

| Gaussian Scene | Scene PSNR | | | | |
| --- | --- | --- | --- | --- | --- |
| | Waldo kitchen | Figurines | Ramen | Teatime | Mean |
| 3DGS | 32.89 | 24.95 | 28.73 | 31.39 | 29.49 |
| Corr-init | 32.18 | 24.61 | 28.84 | 31.52 | 29.29 |
| Corr-init + Densify | 34.65 | 26.34 | 30.02 | 32.73 | 30.94 |

Table 9: Optimization time analysis on enabling densification for Gaussian optimization of ProFuse pretrained scene.

| Densification | Optimization Time | | | | |
| --- | --- | --- | --- | --- | --- |
| | Waldo kitchen | Figurines | Ramen | Teatime | Mean |
| W/O Densify | ~13m | ~12m | ~14m | ~15m | ~14m |
| Densify | ~27m | ~23m | ~16m | ~24m | ~22m |

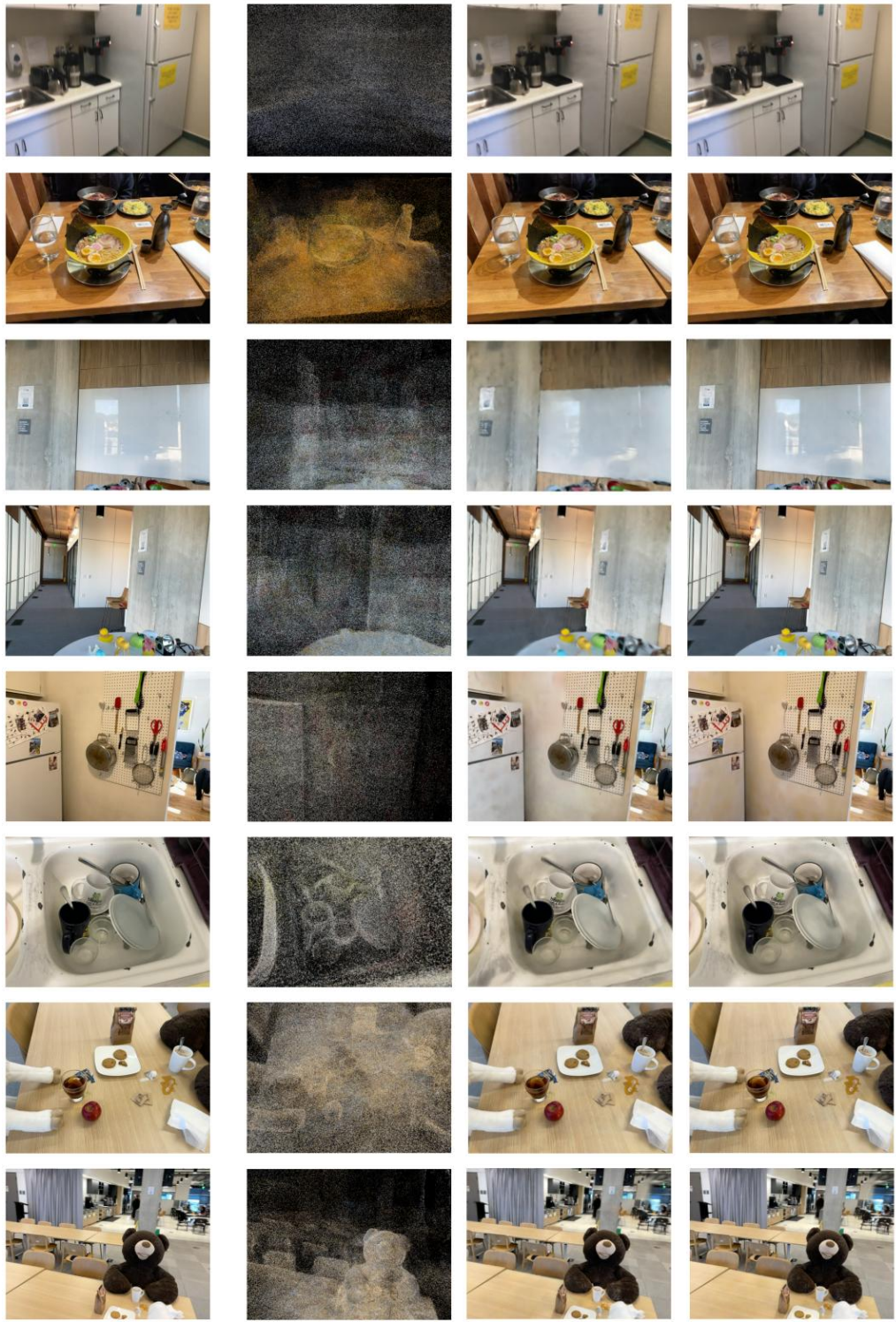

Figure 1. Correspondence–based scene reconstruction of ProFuse.

## E.2. Neighbor per Reference

We study how the number of neighbors per reference view influences both semantic understanding and the cost of dense correspondence. The experiment varies the neighbor count in pre-registration stage, using values 0, 3, 5, 7, and 9 while keeping all other components fixed. Setting the neighbor count to 0 disables cross-view mask clustering and therefore removes 3D Context Proposals, leaving solely correspondence-guided Gaussian initialization. The remaining settings preserve both the correspondence-based initialization and the context proposal pipeline, with increasingly large neighborhood graphs.

Tables 11 and 12 report scene-level mIoU and mAcc@0.25 on the LERF object selection task. Comparing 0 and 3 neighbors shows that introducing even a small neighborhood already improves mIoU noticeably, which confirms that the gain of ProFuse over the registration baseline is not explained only by a different pretrained Gaussian scene. The additional global feature injected by 3D context proposals genuinely strengthens semantic association. As the neighbor count grows beyond 3, several scenes exhibit further improvements in mIoU, while the Figurines scene becomes slightly less stable, suggesting that very large neighborhoods may introduce noisy cross-view links for cluttered layouts. Overall increase in mIoU indicates that richer cross-view evidence can still benefit the registration stage.

The mAcc@0.25 curves show a more nuanced behavior. Waldo kitchen is a representative example. Its mIoU increases significantly when the neighbor count exceeds 3, yet its mAcc decreases slightly. This apparent contradiction is explained by the threshold-selection procedure. For each variant we fix a single activation threshold shared across scenes. The mIoU of Waldo kitchen as a function of the threshold forms a curve whose peak shifts toward lower thresholds when the neighborhood grows. The global threshold chosen for the ablation lies farther from this new peak, so the reported mAcc does not fully reflect the best possible accuracy of the scene. In practice Waldo kitchen can reach mIoU above 46 and mAcc above 68 under a threshold tuned specifically for that configuration, which is consistent with the improved curve.

Table 13 summarizes the initialization time of the dense correspondence stage. The cost grows steadily with the neighbor count, since every additional neighbor requires extra RoMa evaluation and mask projection. Different scenes exhibit slightly different sensitivity, but the trend is consistent. When we compare the average semantic gains against the additional time, three neighbors per reference offers a favorable trade-off, keeping the pre-registration stage within a practical budget.

Table 10: Ablation study on scene mIoU for different neighbor per reference.

| Neighbors | Scene mIoU | | | | |
| --- | --- | --- | --- | --- | --- |
| | Waldo kitchen | Figurines | Ramen | Teatime | Mean |
| 0 | 33.80 | 48.28 | 25.09 | 57.57 | 41.19 |
| 3 | 36.91 | 56.13 | 28.16 | 62.78 | 46.00 |
| 5 | 40.89 | 50.06 | 29.29 | 64.99 | 46.31 |
| 7 | 43.08 | 49.31 | 28.99 | 65.10 | 46.62 |
| 9 | 43.51 | 49.15 | 29.10 | 65.04 | 46.7 |

Table 11: Ablation study on scene mAcc for different neighbor per reference.

| Neighbors | Scene mAcc@0.25 | | | | |
| --- | --- | --- | --- | --- | --- |
| | Waldo kitchen | Figurines | Ramen | Teatime | Mean |
| 0 | 63.64 | 82.14 | 38.03 | 77.97 | 64.45 |
| 3 | 68.18 | 85.71 | 39.44 | 79.66 | 68.25 |
| 5 | 50.00 | 83.93 | 40.85 | 79.66 | 63.61 |
| 7 | 54.54 | 82.14 | 40.85 | 79.66 | 64.30 |
| 9 | 54.54 | 82.14 | 40.85 | 79.66 | 64.30 |

Table 12: Ablation Study on Dense correspondence initialization time with different neighbor per reference

| Neighbors | Scene Init Time | | | | |
| --- | --- | --- | --- | --- | --- |
| | Waldo kitchen | Figurines | Ramen | Teatime | Mean |
| 3 | 2m 29s | 2m 37s | 1m 40s | 1m 47s | 2m 08s |
| 5 | 4m 1s | 3m 43s | 3m 18s | 3m 51s | 3m 44s |
| 7 | 3m 58s | 6m 21s | 3m 47s | 5m 35s | 4m 25s |
| 9 | 7m 11s | 8m 4s | 4m 52s | 6m 53s | 6m 23s |

## E.3. Additional Ablations

**More Details on Top-K Choice.** Top-K ablations for ScanNet point-cloud understanding with 10 labeled classes were discussed in Section 4.5. The same behavior also applied to 19 and 15 classes label sets. Under the same evaluation protocol, ProFuse reaches its best or near-best mIoU and mAcc within top-10 Gaussians, while the Dr. Splat baseline continues to improve when K is increased and usually needs K=40 to approach its own peak. This contrast indicates that our proposal-based registration concentrates the useful semantic mass on a much smaller subset of Gaussians along each ray and is therefore far more efficient.

We seek to find direct evidence for this concentration and whether the outcome is healthy by analyzing the contribution mass of each scene. For each ScanNet scene, we record the per-view "top-10 share" during registration. For each view, a vector is built where each entry is the total contribution weight assigned to one context proposal in that view. We then derive top 10 share as

$$\frac{mass\ of\ 10\ largest\ proposals}{total\ mass\ of\ all\ proposals}.$$

The resulting mass for each view ranges between 0.87 and 0.99, which means that the 10 most active proposals already account for the majority of the total proposal mass in a typical view. At the scene level, we also compare the fraction of total Gaussian mass that lies in the most heavily used 0.1% and 1% of Gaussians. The top 1% of Gaussians carries between 41 ~ 68% of the accumulated contribution mass, and the top 0.1% still carry 14 ~ 42%. These statistics show that both clusters and Gaussians exhibit a highly skewed distribution under our registration scheme, which explains why ProFuse saturates at K=10 on all three ScanNet label sets, whereas Dr. Splat requires much larger K to reach comparable point-cloud performance.

**Warped IoU.** We ablate the warped intersection-over-union threshold $\tau_{iou}$ that decides whether a dense-warped mask pair contributes an edge to the cluster graph. The threshold is varied in {0.1,0.2,0.3,0.4,0.5} and for each value we recompute 3D context proposals and repeat the object-selection and point cloud evaluation. The resulting scene-level mIoU curves are not monotonic, and the behavior differs across scenes, which suggests that the effective operating point is shaped by a combination of warped IoU and the mutual best-neighbor rule. Once mutual best neighbors are enforced, many noisy correspondences are removed before the IoU gate is applied, resulting in $\tau_{iou}$ controls mainly a secondary pruning stage. In this regime $\tau_{iou} = 0.2$ provides a stable choice across scenes. It retains enough cross-view links to form reliable context proposals while still discarding clearly inconsistent warps, and we adopt this threshold as default setting for all main experiments.

**Inferencing Efficiency.** Removing densification and relying on correspondence-guided initialization leads to a much more compact Gaussian scene. Table 13 summarized the comparison of average inferencing time and total Gaussians per scene. On ScanNet, Dr. Splat keeps 2.7 times more Gaussians than our method while operating on the same data. Despite this reduction, the semantic and geometric metrics reported in the main paper remain competitive or improved, which indicates that the correspondence seeds and pruning schedule preserve the informative splats.

We also measure the cost of point-cloud inference on ScanNet. Under the same evaluation pipeline, Dr. Splat requires on average about 99 seconds per scene, whereas ProFuse completes the same retrieval in about 59 seconds. The method therefore achieves faster inference together with a substantially smaller Gaussian set, which matches the reduction in Gaussian count and confirms that the correspondence-initialized, non-densified scenes are advantageous both for efficiency and for downstream open-vocabulary understanding.

Table 13: Ablation study on inference efficiency.

| Method | Inference Time/ Total Gaussians |
|---|---|
| Dr. Splat | 99s / 1.27M |
| ProFuse | 59s / 470K |

### E.4. More Qualitative Results

We provide additional qualitative results of ProFuse in this section. Figures 2 and 3 visualize cosine similarity activations for several text queries on LERF scenes. The results illustrate that global features sharpen the response on the queried object, suppress background clutter, and maintain consistent activation within the object extent. Figure 4 presents PCA projections of the per-Gaussian language features on ScanNet and LERF scenes. For each view we display the input image together with the projected features of Dr. Splat and ProFuse. The ProFuse features form more coherent regions that align much better with objects and surfaces and they remain stable across different viewpoints, which supports the improvements observed in the quantitative evaluations. Figure 5 illustrate feature visualization on more ScanNet scenes, demonstrating the strength of ProFuse in context understanding.

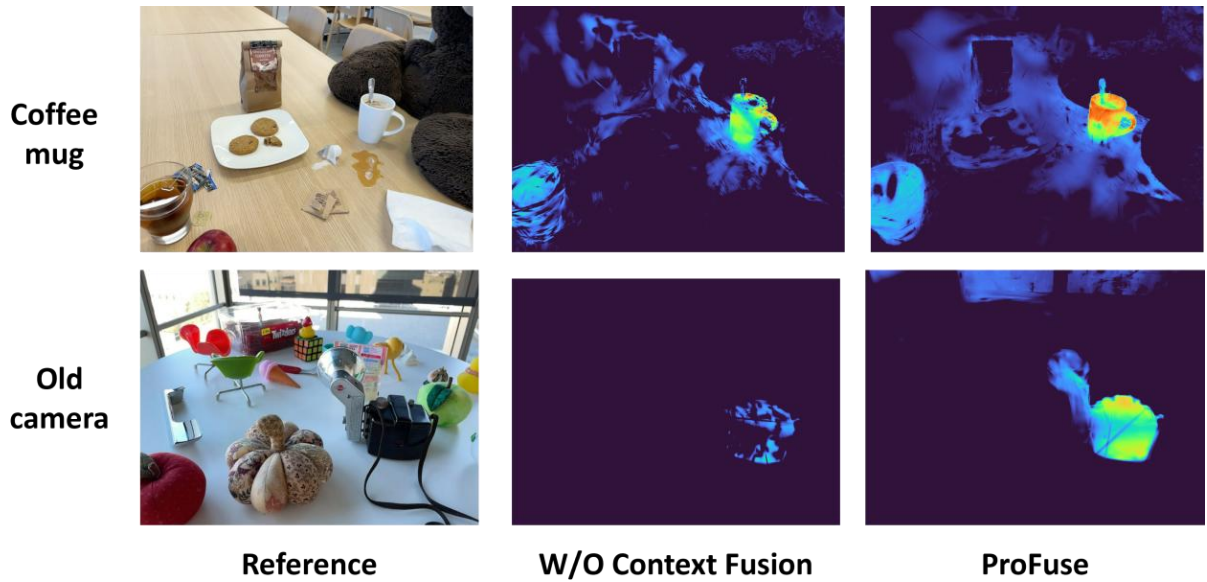

Figure 2. Qualitative results of cosine similarity activation on given queries "coffee mug" and "old camera".

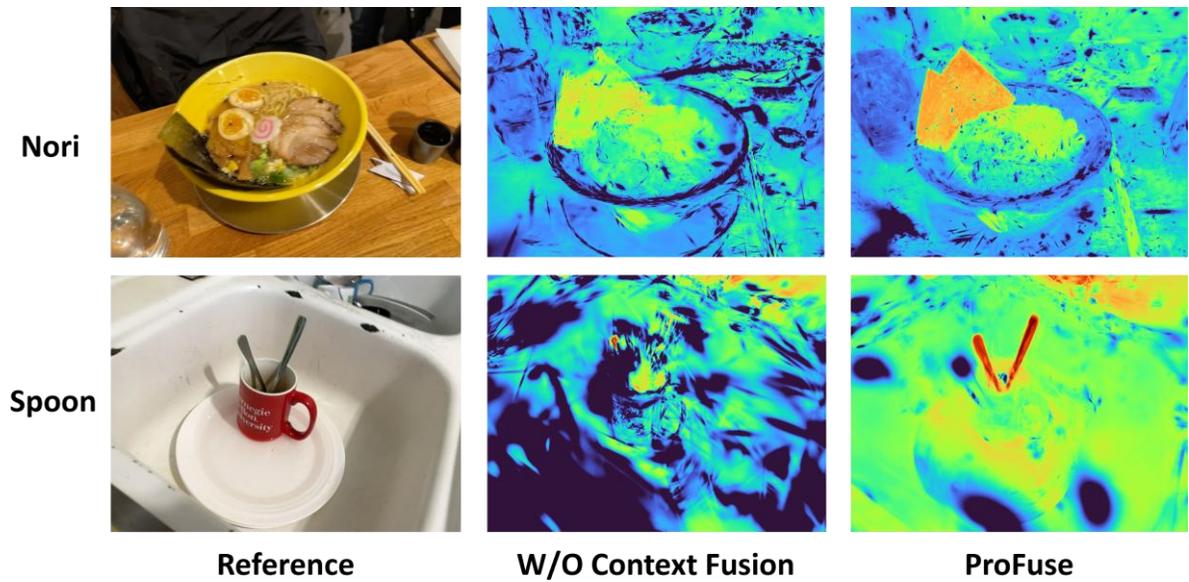

Figure 3. Qualitative results of cosine similarity activation on given queries "nori" and "spoon". With global features, ProFuse carries context-level interpretation and injects consistency in 3D scene understanding.

9

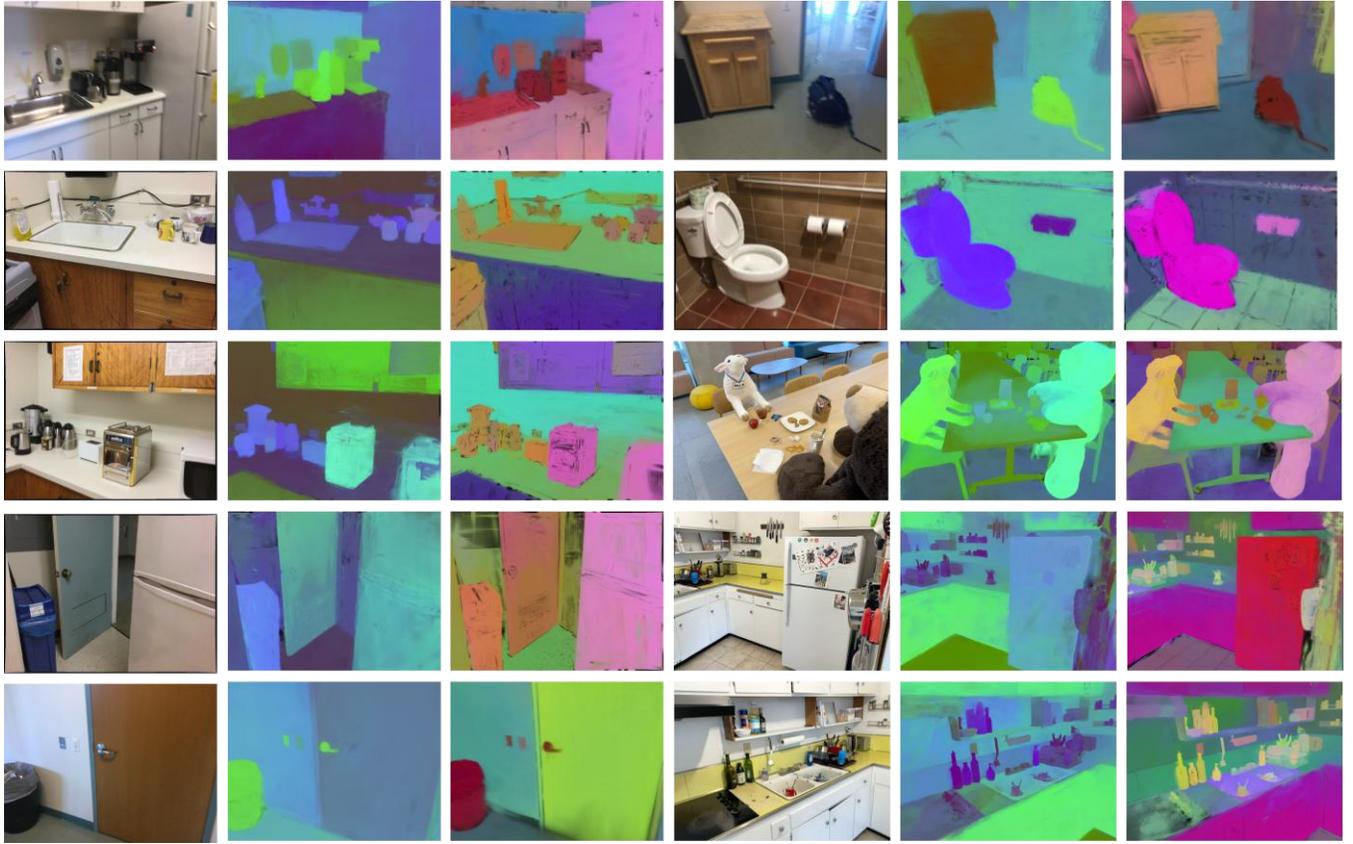

Figure 4. Qualitative results of PCA visualization on ScanNet and LERF scenes. For each view, we provide the reference image (left), and render PCA of Dr. Splat (middle), and ProFuse (right) for comparison.

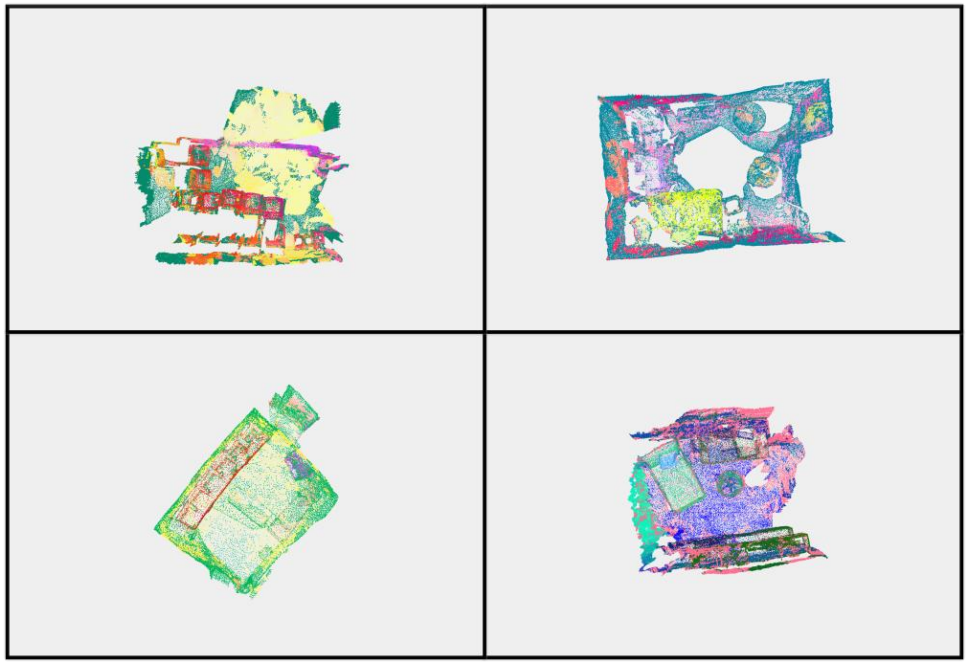

Figure 5. Additional rendering of feature visualization on ScanNet scenes.

10



\end{document}